\documentclass{article}

\usepackage{times}
\usepackage{graphicx} 
\usepackage{subfigure} 
\usepackage{placeins}
\usepackage{natbib}
\usepackage{booktabs}       

\usepackage{algorithm}
\usepackage{algorithmic}

\usepackage{hyperref}

\newcommand{\figref}[1]{Fig.~\ref{#1}}

\renewcommand{\vec}{\mathbf}

\usepackage[accepted]{icml2017}

\icmltitlerunning{Neural Machine Translation in Linear Time}

\begin{document} 

\twocolumn[
\icmltitle{Neural Machine Translation in Linear Time}

\icmlsetsymbol{equal}{*}

\begin{icmlauthorlist} 
\icmlauthor{Nal Kalchbrenner\ }{} 
\icmlauthor{Lasse Espeholt\ }{} 
\icmlauthor{Karen Simonyan\ }{} 
\icmlauthor{A\"aron van den Oord\ }{} 
\icmlauthor{Alex Graves\ }{} 
\icmlauthor{Koray Kavukcuoglu}{}
\icmlauthor{{\normalfont Google Deepmind, London UK}}{} \\
\icmlauthor{{\normalfont\texttt{nalk@google.com}}}{} 
\end{icmlauthorlist}



\icmlkeywords{neural machine translation, convolutional neural networks}

\vskip 0.3in
]

\begin{abstract} 
We present a novel neural network for processing sequences. The ByteNet is a one-dimensional convolutional neural network that is composed of two parts, one to encode the source sequence and the other to decode the target sequence. The two network parts are connected by stacking the decoder on top of the encoder and preserving the temporal resolution of the sequences. To address the differing lengths of the source and the target, we introduce an efficient mechanism by which the decoder is dynamically unfolded over the representation of the encoder. The ByteNet uses dilation in the convolutional layers to increase its receptive field. The resulting network has two core properties: it runs in time that is linear in the length of the sequences and it sidesteps the need for excessive memorization. The ByteNet  decoder attains state-of-the-art performance on character-level language modelling and outperforms the previous best results obtained with recurrent networks. 
The ByteNet also achieves state-of-the-art performance on character-to-character machine translation on the English-to-German WMT translation task, surpassing comparable neural translation models that are based on recurrent networks with attentional pooling and run in quadratic time. We find that the latent alignment structure contained in the representations reflects the expected alignment between the tokens.
\end{abstract} 

\section{Introduction}

In neural language modelling, a neural network estimates a distribution over sequences of words or characters that belong to a given language \citep{bengio:2003}. In neural machine translation, the network estimates a distribution over sequences in the target language conditioned on a given sequence in the source language. The network can be thought of as composed of two parts: a \emph{source network} (the encoder) that encodes the source sequence into a representation and a \emph{target network} (the decoder) that uses the representation of the source encoder to generate the target sequence \citep{kalchbrenner13emnlp}.

\begin{figure}
\centering
\includegraphics[scale=.47]{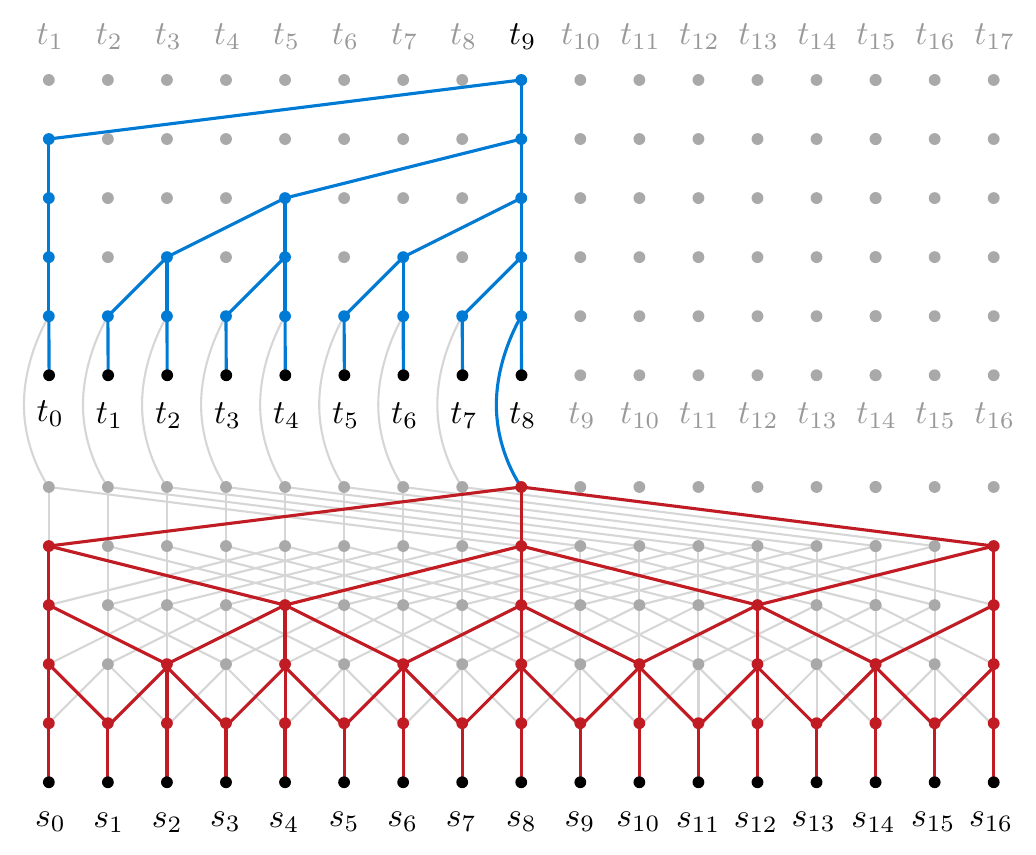}
\caption{The architecture of the ByteNet. The target decoder (blue) is stacked on top of the source encoder (red). The decoder generates the variable-length target sequence using dynamic unfolding.}
\label{architecture}
\end{figure}

Recurrent neural networks (RNN) are powerful sequence models \citep{hochreiter1997long} and are widely used in language modelling \citep{DBLP:conf/interspeech/MikolovKBCK10}, yet they have a potential drawback. RNNs have an inherently serial structure that prevents them from being run in parallel along the sequence length during training and evaluation. Forward and backward signals in a RNN also need to traverse the full distance of the serial path to reach from one token in the sequence to another. The larger the distance, the harder it is to learn the dependencies between the tokens \citep{chapter-gradient-flow-2001}.

\begin{figure*}
\centering
\includegraphics[width=0.8\linewidth]{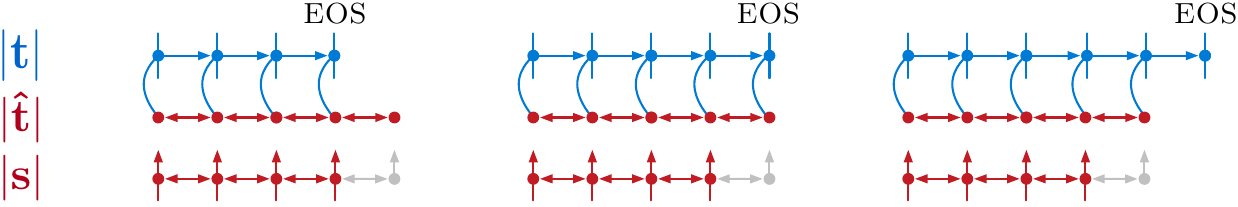}
\caption{Dynamic unfolding in the ByteNet architecture. At each step the decoder is conditioned on the source representation produced by the encoder for that step, or simply on no representation for steps beyond the extended length $|\hat{\vec{t}}|$. The decoding ends when the target network produces an end-of-sequence (EOS) symbol.}
\label{fig:dynunf}
\end{figure*}

A number of neural architectures have been proposed for modelling translation, such as encoder-decoder networks \citep{kalchbrenner13emnlp, DBLP:conf/nips/SutskeverVL14, DBLP:journals/corr/ChoMGBSB14,kaiser2016active}, networks with attentional pooling \citep{DBLP:journals/corr/BahdanauCB14} and two-dimensional networks \citep{DBLP:journals/corr/KalchbrennerDG15}. Despite the generally good performance, the proposed models either have running time that is super-linear in the length of the source and target sequences, or they process the source sequence into a constant size representation, burdening the model with a memorization step. Both of these drawbacks grow more severe as the length of the sequences increases.

We present a family of encoder-decoder neural networks that are characterized by two architectural mechanisms aimed to address the drawbacks of the conventional approaches mentioned above. The first mechanism involves the \emph{stacking} of the decoder on top of the representation of the encoder in a manner that preserves the temporal resolution of the sequences; this is in contrast with architectures that encode the source into a fixed-size representation \citep{kalchbrenner13emnlp, DBLP:conf/nips/SutskeverVL14}.  The second mechanism is the \emph{dynamic unfolding} mechanism that allows the network to process in a simple and efficient way source and target sequences of different lengths (Sect.~\ref{sec:dynunf}).

The ByteNet is the instance within this family of models that uses one-dimensional convolutional neural networks (CNN) of fixed depth for both the encoder and the decoder (\figref{architecture}). The two CNNs use increasing factors of dilation to rapidly grow the receptive fields; a similar technique is also used in \citep{wavenet}. The convolutions in the decoder CNN are masked to prevent the network from seeing future tokens in the target sequence \citep{van2016pixel}.

The network has  beneficial computational and learning properties. From a computational perspective, the network has a running time that is \emph{linear} in the length of the source and target sequences (up to a constant $c \approx \log d$ where $d$ is the size of the desired dependency field). The computation in the encoder during training and decoding and in the decoder during training can also be run efficiently \emph{in parallel} along the sequences (Sect.~\ref{NTM}).
From a learning perspective, the representation of the source sequence in the ByteNet is \emph{resolution preserving}; the representation sidesteps the need for memorization and allows for maximal bandwidth between encoder and decoder. In addition, the distance traversed by forward and backward signals between any input and output tokens corresponds to the fixed depth of the networks and is largely independent of the distance between the tokens. Dependencies over large distances are connected by short paths and can be learnt more easily.

We apply the ByteNet model to strings of characters for character-level language modelling and character-to-character machine translation. We evaluate the decoder network on the Hutter Prize Wikipedia task \citep{hutterprize} where it achieves the state-of-the-art performance of 1.31 bits/character. We further evaluate the encoder-decoder network on character-to-character machine translation on the English-to-German WMT benchmark where it achieves a state-of-the-art BLEU score of 22.85 (0.380 bits/character) and 25.53 (0.389 bits/character) on the 2014 and 2015 test sets, respectively. On the character-level machine translation task, ByteNet betters a comparable version of GNMT \citep{wu2016} that is a state-of-the-art system. These results show that deep CNNs are simple, scalable and effective architectures for challenging linguistic processing tasks.

The paper is organized as follows. Section 2 lays out the background and some desiderata for neural architectures underlying translation models. Section 3 defines the proposed family of architectures and the specific convolutional instance (ByteNet) used in the experiments. Section 4 analyses ByteNet as well as existing neural translation models based on the desiderata set out in Section 2. Section 5 reports the experiments on language modelling and Section 6 reports the experiments on character-to-character machine translation.

\section{Neural Translation Model}
\label{NTM}
Given a string $\vec{s}$ from a source language, a neural translation model estimates a distribution $p(\vec{t}|\vec{s})$ over strings $\vec{t}$ of a target language. The distribution indicates the probability of a string $\vec{t}$ being a translation of $\vec{s}$. A product of conditionals over the tokens in the target $\vec{t} = t_0,...,t_N$ leads to a tractable formulation of the distribution:
\begin{equation}
p(\vec{t} | \vec{s}) = \prod_{i=0}^{N}p(t_{i} | t_{<i},\vec{s})
\label{eq:likelihood}
\end{equation}
Each conditional factor expresses complex and long-range dependencies among the source and target tokens. The strings are usually sentences of the respective languages; the tokens are words or, as in the our case, characters.
The network that models $p(\vec{t}|\vec{s})$ is composed of two parts: a source network (the encoder) that processes the source string into a representation and a target network (the decoder) that uses the source representation to generate the target string \citep{kalchbrenner13emnlp}. The decoder  functions as a language model for the target language.

A neural translation model has some basic properties. The decoder is autoregressive in the target tokens and the model is sensitive to the ordering of the tokens in the source and target strings. It is also useful for the model to be able to assign a non-zero probability to any string in the target language and retain an open vocabulary.

\subsection{Desiderata}
\label{desiderata}

Beyond these basic properties the definition of a neural translation model does not determine a unique neural architecture, so we aim at identifying some desiderata.

First, the running time of the network should be \emph{linear} in the length of the source and target strings. 
This ensures that the model is scalable to longer strings, which is the case when using characters as tokens.

The use of operations that run \emph{in parallel} along the sequence length can also be beneficial for reducing computation time.

Second, the size of the source representation should be linear in the length of the source string, i.e. it should be \emph{resolution preserving}, and not have constant size. This is to avoid burdening the model with an additional memorization step before translation. In more general terms, the size of a representation should be proportional to the amount of information it represents or predicts. 

Third, the path traversed by forward and backward signals in the network (between input and ouput tokens) should be short.
Shorter paths whose length is largely decoupled from the sequence distance between the two tokens have the potential to better propagate the signals \citep{chapter-gradient-flow-2001} and to let the network learn long-range dependencies more easily.

\begin{figure}[t]
    \centering
    \begin{subfigure}{}
        \includegraphics[scale=.28]{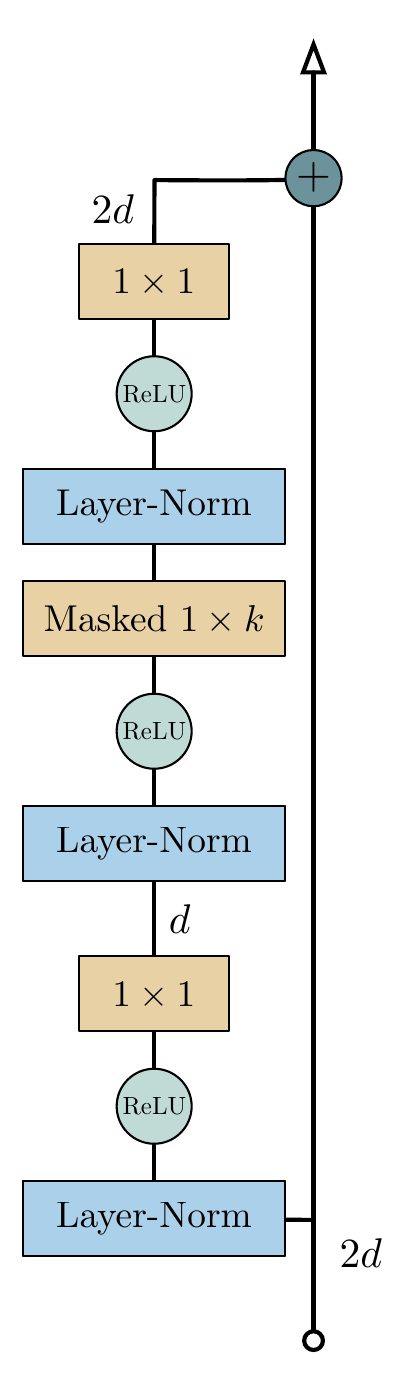}
    \end{subfigure}%
    \hspace{1cm}
    \begin{subfigure}{}
        \includegraphics[scale=.28]{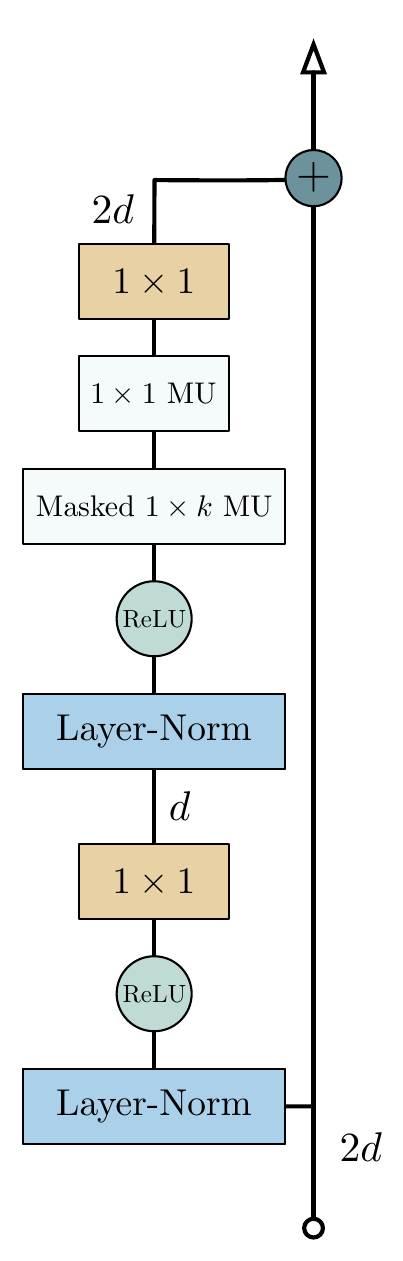}
        \hspace{.25cm}
        \includegraphics[scale=.28]{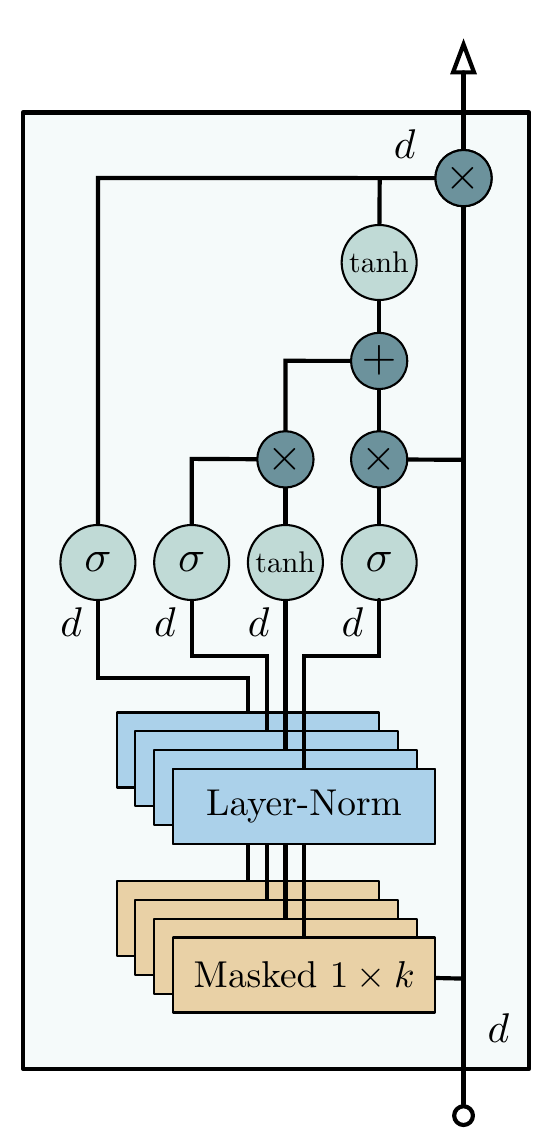}
    \end{subfigure}
    \caption{Left: Residual block with ReLUs \citep{DBLP:journals/corr/HeZR016} adapted for decoders. Right: Residual Multiplicative Block adapted for decoders and corresponding expansion of the MU \citep{vpn}.}
    \label{fig:residual}
\vspace{-0.5cm}
\end{figure}

\section{ByteNet}
\label{bytenet}
We aim at building neural language and translation models that capture the desiderata set out in Sect.~\ref{desiderata}. The proposed ByteNet architecture is composed of a decoder that is \emph{stacked} on an encoder (Sect.~\ref{sec:stacking}) and generates variable-length outputs via \emph{dynamic unfolding} (Sect.~\ref{sec:dynunf}). The decoder is a language model that is formed of one-dimensional convolutional layers that are masked (Sect.~\ref{masked}) and use dilation (Sect.~\ref{dilation}).  The encoder processes the source string into a representation and is formed of one-dimensional convolutional layers that use dilation but are \emph{not} masked. Figure~\ref{architecture} depicts the two networks and their combination.

\subsection{Encoder-Decoder Stacking}
\label{sec:stacking}
A notable feature of the proposed family of architectures is the way the encoder and the decoder are connected. To maximize the representational bandwidth between the encoder and the decoder, we place the decoder on top of the representation computed by the encoder. This is in contrast to models that compress the source representation into a fixed-size vector \citep{kalchbrenner13emnlp,DBLP:conf/nips/SutskeverVL14} or that pool over the source representation with a mechanism such as attentional pooling \citep{DBLP:journals/corr/BahdanauCB14}. 

\subsection{Dynamic Unfolding}
\label{sec:dynunf}
An encoder and a decoder network that process sequences of different lengths cannot be directly connected due to the different sizes of the computed representations. We circumvent this issue via a mechanism which we call dynamic unfolding, which works as follows.

\begin{figure}[t]
    \centering
    \includegraphics[scale=.205, trim={0.1cm 0 0 0},clip]{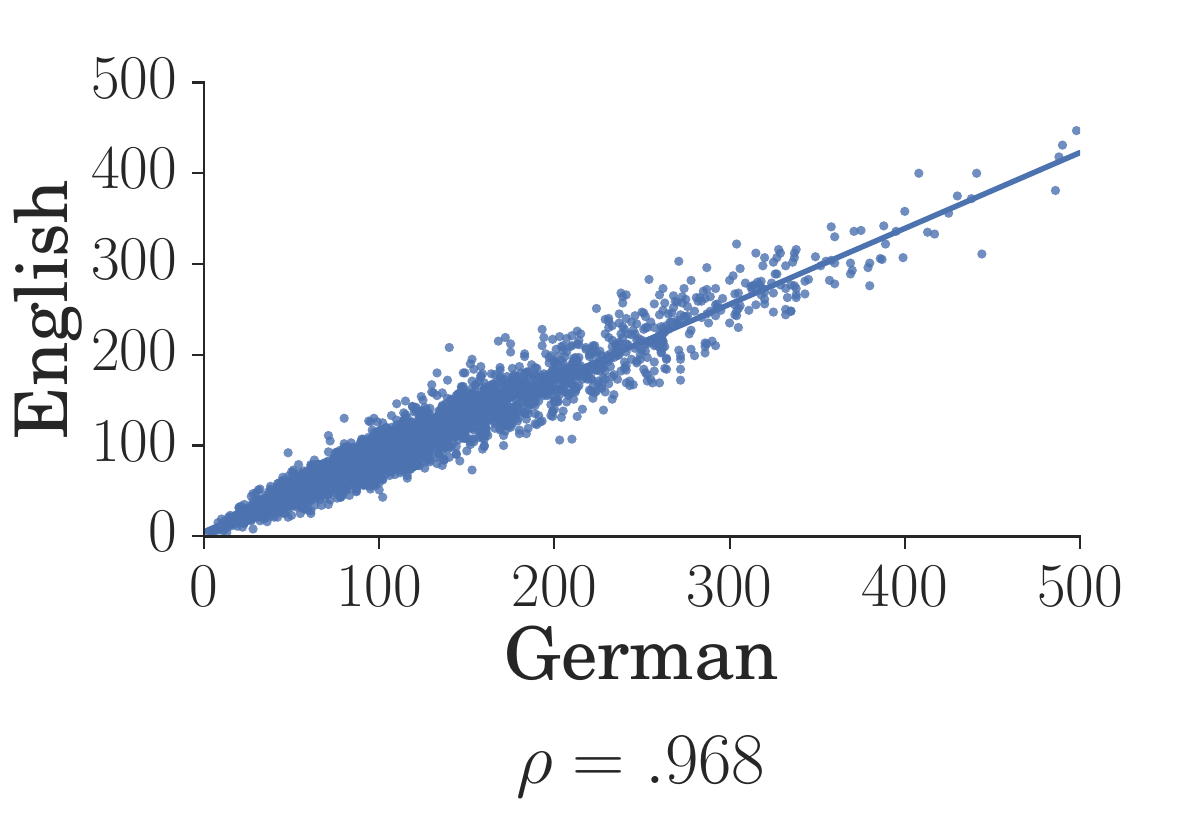}
    \caption{Lengths of sentences in characters and their correlation coefficient for the English-to-German WMT NewsTest-2013 validation data. The correlation coefficient is similarly high ($\rho>0.96$) for all other language pairs that we inspected.}
\label{fig:corr}
\end{figure}

Given source and target sequences $\vec{s}$ and $\vec{t}$ with respective lengths $|\vec{s}|$ and $|\vec{t}|$, one first chooses a sufficiently tight upper bound $\hat{|\vec{t}|}$ on the target length $|\vec{t}|$ as a linear function of the source length $|\vec{s}|$:
\begin{equation}
\label{eq:unfold}
\hat{|\vec{t}|} = a |\vec{s}| + b
\end{equation}
The tight upper bound $\hat{|\vec{t}|}$ is chosen in such a way that, on the one hand, it is greater than the actual length $|\vec{t}|$ in almost all cases and, on the other hand, it does not increase excessively the  amount of computation that is required. Once a linear relationship is chosen, one designs the source encoder so that, given a source sequence of length $|\vec{s}|$, the encoder outputs a representation of the established length $\hat{|\vec{t}|}$. In our case, we let $a=1.20$ and $b=0$ when translating from English into German, as German sentences tend to be somewhat longer than their English counterparts (\figref{fig:corr}). In this manner the representation produced by the encoder can be efficiently computed, while maintaining high bandwidth and being resolution-preserving. Once the encoder representation is computed, we let the decoder unfold step-by-step over the encoder representation until the decoder itself outputs an end-of-sequence symbol; the unfolding process may freely proceed beyond the estimated length $\hat{|\vec{t}|}$ of the encoder representation. Figure~\ref{fig:dynunf} gives an example of dynamic unfolding.

\subsection{Input Embedding Tensor}
Given the target sequence $\vec{t} = t_0,...,t_{n}$ the ByteNet decoder embeds each of the first $n$ tokens $t_0,...,t_{n-1}$ via a look-up table (the $n$ tokens $t_1,...,t_{n}$ serve as targets for the predictions). The resulting embeddings are concatenated into a tensor of size $n \times 2d$ where $d$ is the number of inner channels in the network.

\subsection{Masked One-dimensional Convolutions}
\label{masked}
The decoder applies masked one-dimensional convolutions \citep{van2016pixel} to the input embedding tensor that have a masked kernel of size $k$. The masking ensures that information from future tokens does not affect the prediction of the current token.  The operation can be implemented either by zeroing out some of the weights of a wider kernel of size $2k-1$ or by padding the input map. 

\begin{figure}[t]
\centering
\hspace{-1cm}
\begin{subfigure}{}
    \includegraphics[scale=0.5]{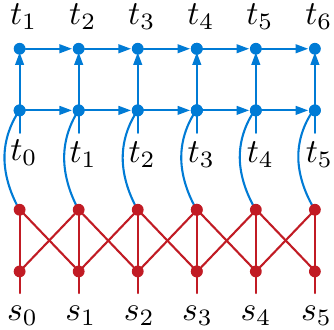}
    \label{fig:mt_decoder_block}
\end{subfigure}
\hspace{1cm}
\begin{subfigure}{}
    \includegraphics[scale=0.5]{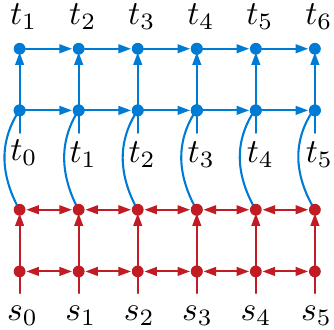}
    \label{fig:lm_decoder_block}
\end{subfigure}
\hspace{-1cm}
\caption{Recurrent ByteNet variants of the ByteNet architecture. Left: Recurrent ByteNet with convolutional source network and recurrent target network. Right: Recurrent ByteNet with bidirectional recurrent source network and recurrent target network. The latter architecture is a strict generalization of the RNN Enc-Dec network. }
\label{recurrentBytenet}
\end{figure}

\subsection{Dilation}
\label{dilation}

The masked convolutions use dilation to increase the receptive field of the target network~\citep{DBLP:journals/corr/ChenPKMY14,DBLP:journals/corr/YuK15}. Dilation makes the receptive field grow exponentially in terms of the depth of the networks, as opposed to linearly.  We use a dilation scheme whereby the dilation rates are doubled every layer up to a maximum rate $r$ (for our experiments $r=16$). The scheme is repeated multiple times in the network always starting from a dilation rate of 1~\citep{wavenet,vpn}. 

\subsection{Residual Blocks}

Each layer is wrapped in a residual block that contains additional convolutional layers with filters of size $1\times1$ \citep{DBLP:journals/corr/HeZR016}. We adopt two variants of the residual blocks: one with ReLUs, which is used in the machine translation experiments, and one with Multiplicative Units \citep{vpn}, which is used in the language modelling experiments. Figure~\ref{fig:residual} diagrams the two variants of the blocks. In both cases, we use layer normalization~\citep{DBLP:journals/corr/BaKH16} before the activation function, as it is well suited to sequence processing where computing the activation statistics over the following future tokens (as would be done by batch normalization) must be avoided.
After a series of residual blocks of increased dilation, the network applies one more convolution  and ReLU followed by a convolution and a final softmax layer.

\section{Model Comparison}
\label{modelcomp}

In this section we analyze the properties of various previously introduced neural translation models as well as the ByteNet family of models. For the sake of a more complete analysis, we include two recurrent ByteNet variants (which we do not evaluate in the experiments).

\begin{table*}
\small
    \centering
    \bgroup
    \def\arraystretch{1.5}
    \begin{tabular}{l c  c c c c c}
    
        \toprule
        
       \textbf{Model}  & \textbf{Net$_\mathbf{S}$} & \textbf{Net$_\mathbf{T}$} & \textbf{Time} & \textbf{RP} & \textbf{Path$_\mathbf{S}$} & \textbf{Path$_\mathbf{T}$} \\
        \hline
        \multicolumn{1}{l}{RCTM 1 } & CNN & RNN & $|S||S|+|T|$ & no & $|S|$  & $|T|$  \\ 
        \multicolumn{1}{l}{RCTM 2 } & CNN & RNN & $|S||S| + |T|$ & yes & $|S|$  & $|T|$ \\
        \multicolumn{1}{l}{RNN Enc-Dec } & RNN & RNN & $|S| + |T|$ & no & $|S|+|T|$ & $|T|$  \\
        \multicolumn{1}{l}{RNN Enc-Dec Att  } & RNN & RNN & $|S||T|$ &
         yes & 1 &  $|T|$ \\
        \multicolumn{1}{l}{Grid LSTM } & RNN & RNN & $|S||T|$ &
        yes & $|S| + |T|$ & $|S|+|T|$ \\ 
        \multicolumn{1}{l}{Extended Neural GPU } & cRNN & cRNN & $|S||S| + |S||T|$ &
        yes & $|S|$ & $|T|$  \\ \hline
        \multicolumn{1}{l}{Recurrent ByteNet} & RNN & RNN & $|S|+|T|$ & yes
         & $\max(|S|,|T|)$ & $|T|$  \\
        \multicolumn{1}{l}{Recurrent ByteNet} & CNN & RNN & $c|S|+|T|$ &
        yes & $c$ & $|T|$  \\
        \multicolumn{1}{l}{ByteNet} & CNN & CNN & $c|S|+c|T|$ &
        yes & $c$ & $c$ \\ \bottomrule
    \end{tabular}
    \egroup
    \caption{ Properties of various neural translation models.}
\label{properties}
\end{table*}

\subsection{Recurrent ByteNets}
The ByteNet is composed of two stacked encoder and decoder networks where the decoder network dynamically adapts to the output length. This way of combining the networks is not tied to the networks being strictly convolutional. We may consider two variants of the ByteNet that use recurrent networks for one or both of the networks (see Figure~\ref{recurrentBytenet}).
The first variant replaces the convolutional decoder with a recurrent one that is similarly stacked and dynamically unfolded. The second variant also replaces the convolutional encoder with a recurrent encoder, e.g.\ a bidirectional RNN. The target RNN is then placed on top of the source RNN. Considering the latter Recurrent ByteNet, we can see that the  RNN Enc-Dec network  \citep{DBLP:conf/nips/SutskeverVL14,DBLP:journals/corr/ChoMGBSB14}  is a Recurrent ByteNet where all connections between source and target -- except for the first one that connects $s_0$ and $t_0$ -- have been severed. The Recurrent ByteNet is a generalization of the RNN Enc-Dec and, modulo the type of weight-sharing scheme, so is the convolutional ByteNet.

\subsection{Comparison of Properties}

In our comparison we consider the following neural translation models: the Recurrent Continuous Translation Model (RCTM) 1 and 2 \citep{kalchbrenner13emnlp}; the RNN Enc-Dec \citep{DBLP:conf/nips/SutskeverVL14,DBLP:journals/corr/ChoMGBSB14}; the RNN Enc-Dec Att with the attentional pooling mechanism \citep{DBLP:journals/corr/BahdanauCB14} of which there are a few variations \citep{luong-pham-manning:2015:EMNLP,chung2016hierarchical}; the Grid LSTM translation model \citep{DBLP:journals/corr/KalchbrennerDG15} that uses a multi-dimensional architecture; the Extended Neural GPU model \citep{kaiser2016active} that has a convolutional RNN architecture; the ByteNet and the two Recurrent ByteNet variants.

Our comparison criteria reflect the desiderata set out in Sect.~\ref{desiderata}.
We separate the first (computation time) desideratum into three columns. The first column indicates the time complexity of the network as a function of the length of the sequences and is denoted by \textbf{Time}. The other two columns \textbf{Net$_\mathbf{S}$} and \textbf{Net$_\mathbf{T}$} indicate, respectively, whether the source and the target network use a convolutional structure (CNN) or a recurrent one (RNN); a CNN structure has the advantage that it can be run in parallel along the length of the sequence. 
The second (resolution preservation) desideratum corresponds to the \textbf{RP} column, which indicates whether the source representation in the network is resolution preserving. 
Finally, the third desideratum (short forward and backward flow paths) is reflected by two columns.
The \textbf{Path$_\mathbf{S}$} column corresponds to the length in layer steps of the shortest path between a source token and any output target token. Similarly, the \textbf{Path$_\mathbf{T}$} column corresponds to the length of the shortest path between an input target token and any output target token. 
Shorter paths lead to better forward and backward signal propagation.

Table~\ref{properties} summarizes the properties of the models. The ByteNet, the Recurrent ByteNets and the RNN Enc-Dec are the only networks that have linear running time (up to the constant $c$). The RNN Enc-Dec, however, does not preserve the source sequence resolution, a feature that aggravates learning for long sequences such as those that appear in character-to-character machine translation  \citep{DBLP:conf/acl/LuongM16}. The RCTM 2, the RNN Enc-Dec Att, the Grid LSTM and the Extended Neural GPU do preserve the resolution, but at a cost of a quadratic running time. The ByteNet stands out also for its \textbf{Path} properties. The dilated structure of the convolutions connects any two source or target tokens in the sequences by way of a small number of network layers corresponding to the depth of the source or target networks. For character sequences where learning long-range dependencies is important, paths that are sub-linear in the distance are advantageous.

\begin{table*}[t]
\small
  \begin{center}
  \begin{tabular}{lcccc}
    \toprule
    \textbf{Model}  & \textbf{Inputs} & \textbf{Outputs} & \textbf{WMT Test '14} & \textbf{WMT Test '15} \\ \midrule
      Phrase Based MT \citep{freitag14:wmtEuBridge, DBLP:conf/wmt/WilliamsSNHK15} & phrases & phrases & ${20.7}$ & $24.0$ \\ \midrule
      RNN Enc-Dec \citep{luong-pham-manning:2015:EMNLP} & words & words & $11.3$   \\
      Reverse RNN Enc-Dec \citep{luong-pham-manning:2015:EMNLP} & words & words & $14.0$   \\
      RNN Enc-Dec Att \cite{zhou2016deep} & words & words & $20.6$ \\
      RNN Enc-Dec Att \citep{luong-pham-manning:2015:EMNLP} & words & words & ${20.9}$ \\
        GNMT   (RNN Enc-Dec Att) \citep{wu2016} & word-pieces & word-pieces &  $\mathbf{24.61}$ &  \\ \midrule
      RNN Enc-Dec Att \citep{DBLP:conf/acl/ChungCB16} & BPE & BPE & ${19.98}$ & ${21.72}$ \\

      RNN Enc-Dec Att \citep{DBLP:conf/acl/ChungCB16}  & BPE & char & ${21.33}$ & ${23.45}$ \\
      GNMT (RNN Enc-Dec Att) \citep{wu2016} & char & char & $22.62$ \\

      \textbf{ByteNet} & char & char &  $\mathbf{23.75}$ & $\mathbf{26.26}$ \\ 

      \bottomrule
  \end{tabular}
  \end{center}
\caption{BLEU scores on En-De WMT NewsTest 2014 and 2015 test sets.}
\label{mt}
\end{table*}

\begin{table}[t]
\small
  \begin{center}
  \begin{tabular}{l@{\hspace{.22cm}}c@{\hspace{.22cm}}}
    \toprule
    \textbf{Model} &  \textbf{Test}  \\ \midrule
      Stacked LSTM \citep{DBLP:journals/corr/Graves13} & 1.67 \\
      GF-LSTM \citep{chung2015gated} & 1.58 \\ 
      Grid-LSTM \citep{DBLP:journals/corr/KalchbrennerDG15} & 1.47 \\ 
      Layer-normalized LSTM \citep{chung2016hierarchical} & 1.46  \\
      MI-LSTM \citep{wu2016multiplicative} & 1.44 \\
      Recurrent Memory Array Structures \citep{rocki2016recurrent} & 1.40 \\
      HM-LSTM \citep{chung2016hierarchical} & 1.40 \\
      Layer Norm HyperLSTM \citep{2016arXiv160909106H} & 1.38 \\
      Large Layer Norm HyperLSTM \citep{2016arXiv160909106H} & 1.34 \\
      Recurrent Highway Networks \citep{DBLP:journals/corr/SrivastavaGS15}& 1.32 \\
      \textbf{ByteNet Decoder} &  $\mathbf{1.31}$ \\
      \bottomrule
  \end{tabular}
  \end{center}
\caption{Negative log-likelihood results in bits/byte on the Hutter Prize Wikipedia benchmark.}
\label{wiki}
\end{table}

\section{Character Prediction}

We first evaluate the ByteNet Decoder separately on a character-level language modelling benchmark. We use the Hutter Prize version of the Wikipedia dataset and follow the standard split where the first 90 million bytes are used for training, the next 5 million bytes are used for validation and the last 5 million bytes are used for testing \citep{chung2015gated}. The total number of characters in the vocabulary is 205. 

The ByteNet Decoder that we use for the result has 30 residual blocks split into six sets of five blocks each; for the five blocks in each set the dilation rates are, respectively, $1,2,4,8$ and $16$. The masked kernel has size 3. This gives a receptive field of 315 characters. The number of hidden units $d$ is 512.  For this task we use residual multiplicative blocks  (\figref{fig:residual} Right).
For the optimization we use Adam \citep{DBLP:journals/corr/KingmaB14} with a learning rate of $0.0003$ and a weight decay term of $0.0001$. We apply dropout to the last ReLU layer before the softmax dropping units with a probability of 0.1. We do not reduce the learning rate during training. At each step we sample a batch of sequences of 500 characters each, use the first 100 characters as the minimum context and predict the latter 400 characters.

Table~\ref{wiki} lists recent results of various neural sequence models on the Wikipedia dataset. All the results except for the ByteNet result are obtained using some variant of the LSTM recurrent neural network \citep{hochreiter1997long}. The ByteNet decoder achieves 1.31 bits/character on the test set.

\begin{table}[t]
\small
  \begin{center}
  \begin{tabular}{lcc}
    \toprule
      & \textbf{WMT Test '14} & \textbf{WMT Test '15} \\ \midrule
      Bits/character & 0.521 & 0.532  \\ \midrule
      BLEU & 23.75 & 26.26 \\
      \bottomrule
  \end{tabular}
  \end{center}
\caption{Bits/character with respective BLEU score achieved by the ByteNet translation model on the English-to-German WMT translation task.}
\label{nllresults}
\end{table}

\begin{table*}[t]
\small
  \begin{center}
  \begin{tabular}{lccc}
    \toprule
    \textit{Director Jon Favreau, who is currently working on Disney's forthcoming Jungle Book film,} \\ \textit{told the website Hollywood Reporter: ``I think times are changing."}  \vspace{0.2cm} \\ 
    \textit{Regisseur Jon Favreau, der derzeit an Disneys bald erscheinenden Dschungelbuch-Film arbeitet,} \\ \textit{sagte gegenüber der Webseite Hollywood Reporter: ``Ich glaube, die Zeiten \"andern sich."}  \vspace{0.2cm} \\ 
    \textit{Regisseur Jon Favreau, der zur Zeit an Disneys kommendem Jungle Book Film arbeitet,} \\ \textit{ hat der Website Hollywood Reporter gesagt: ``Ich denke, die Zeiten \"andern sich".}  \\
    \midrule
    \textit{Matt Casaday, 25, a senior at Brigham Young University, says he had paid 42 cents on Amazon.com} \\ \textit{for a used copy of ``Strategic Media Decisions: Understanding The Business End Of The Advertising Business."} \vspace{0.2cm} \\
    \textit{Matt Casaday, 25, Abschlussstudent an der Brigham Young University, sagt, dass er auf Amazon.com 42 Cents ausgegeben hat} \\ \textit{f\"ur eine gebrauchte Ausgabe von ``Strategic Media Decisions: Understanding The Business End Of The Advertising Business."} \vspace{0.2cm}
    \\
    \textit{Matt Casaday, 25, ein Senior an der Brigham Young University, sagte, er habe 42 Cent auf Amazon.com} \\ \textit{f\"ur eine gebrauchte Kopie von ``Strategic Media Decisions: Understanding The Business End Of The Advertising Business".}  \\ 
      \bottomrule
  \end{tabular}
  \end{center}
\caption{Raw output translations generated from the ByteNet that highlight interesting reordering and transliteration phenomena. For each group, the first row is the English source, the second row is the ground truth German target, and the third row is the ByteNet translation.}
\label{tab:samples}
\end{table*}

\section{Character-Level Machine Translation}

We evaluate the full ByteNet on the WMT English to German translation task. We use NewsTest 2013 for validation and NewsTest 2014 and 2015 for testing. The English and German strings are encoded as sequences of characters; no explicit segmentation into words or morphemes is applied to the strings. The outputs of the network are strings of characters in the target language. We keep 323 characters in the German vocabulary and 296 in the English vocabulary.

The ByteNet used in the experiments has 30 residual blocks in the encoder and 30 residual blocks in the decoder. As in the ByteNet Decoder, the residual blocks are arranged in sets of five with corresponding dilation rates of $1,2,4,8$ and $16$. For this task we use the residual blocks with ReLUs (\figref{fig:residual} Left). The number of hidden units $d$ is 800. The size of the kernel in the source network is $3$, whereas the size of the masked kernel in the target network is $3$. For the optimization we use Adam with a learning rate of $0.0003$.

Each sentence is padded with special characters to the nearest greater multiple of 50; 20\% of further padding is applied to each source sentence as a part of dynamic unfolding (eq.~\ref{eq:unfold}). Each pair of sentences is mapped to a bucket based on the pair of padded lengths for efficient batching during training. We use \emph{vanilla} beam search according to the total likelihood of the generated candidate and accept only candidates which end in a end-of-sentence token. We use a beam of size 12. We do  not use length normalization, nor do we keep score of which parts of the source sentence have been translated \citep{wu2016}.

Table~\ref{mt} and Table~\ref{nllresults} contain the results of the experiments. On NewsTest 2014 the ByteNet achieves the highest performance in character-level and subword-level neural machine translation, and compared to the word-level systems it is second only to the version of GNMT that uses word-pieces. On NewsTest 2015, to our knowledge, ByteNet achieves the best published results to date.

\begin{figure}
\centering
\includegraphics[width=1\linewidth, trim={0 0 0 2.7cm},clip]{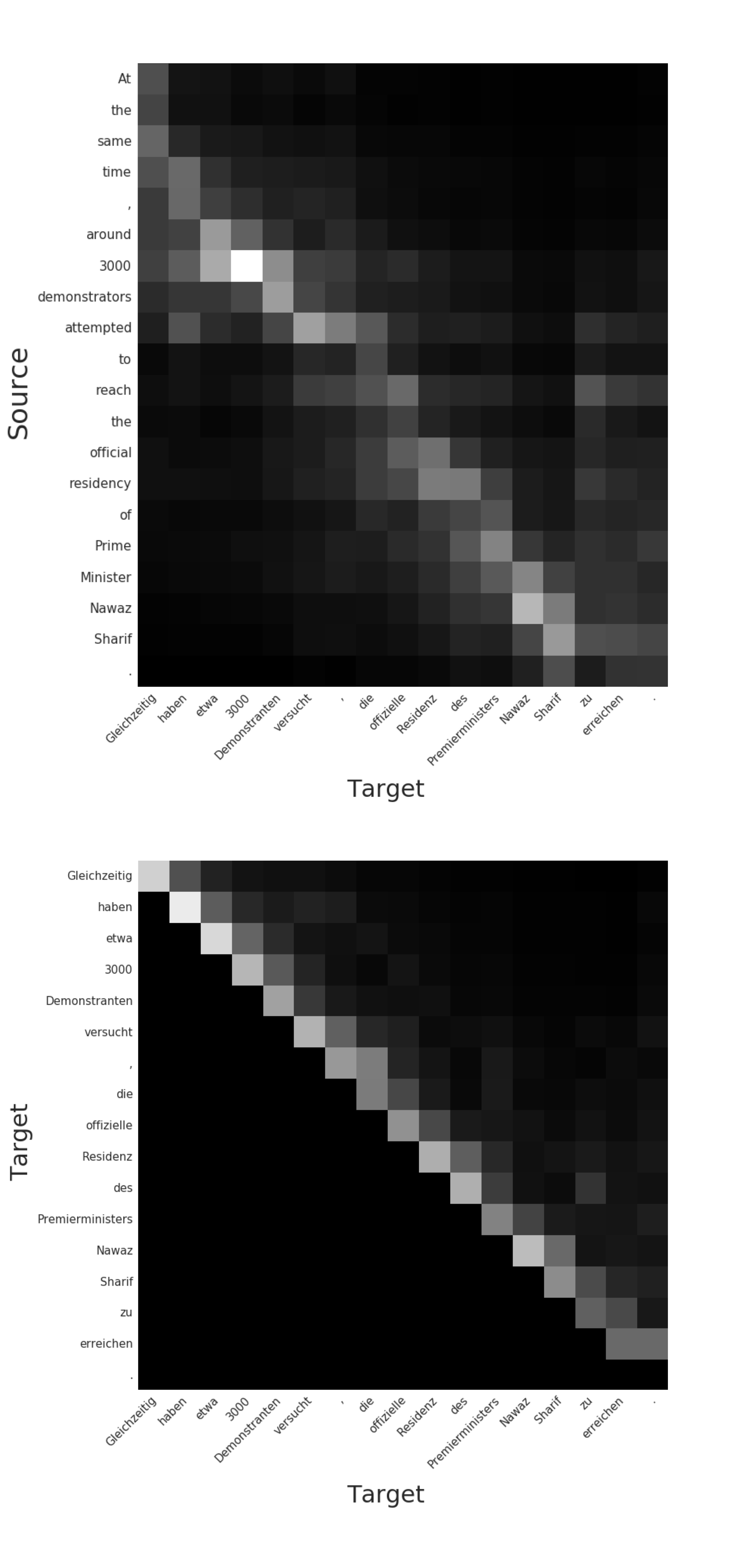}
\vspace{-0.9cm}
\caption{Magnitude of gradients of the predicted outputs with respect to the source and target inputs. The gradients are summed for all the characters in a given word. In the bottom heatmap the magnitudes are nonzero on the diagonal, since the prediction of a target character depends highly on the  preceding target character in the same word.}
\label{fig:gradheatmaps}
\end{figure}

Table~\ref{tab:samples} contains some of the unaltered generated translations from the ByteNet that highlight reordering and other phenomena such as transliteration. The character-level aspect of the model makes post-processing unnecessary in principle. We further visualize the sensitivity of the ByteNet's predictions to specific source and target inputs using gradient-based visualization \citep{DBLP:journals/corr/SimonyanVZ13}. Figure~\ref{fig:gradheatmaps} represents a heatmap of the magnitude of the gradients of the generated outputs with respect to the source and target inputs. For visual clarity, we sum the gradients for all the characters that make up each word and normalize the values along each column. In contrast with the attentional pooling mechanism \citep{DBLP:journals/corr/BahdanauCB14}, this general technique allows us to inspect not just dependencies of the outputs on the source inputs, but also dependencies of the outputs on previous target inputs, or on any other neural network layers.

\section{Conclusion}

We have introduced the ByteNet, a neural translation model that has linear running time, decouples translation from memorization and has short signal propagation paths for tokens in sequences. We have shown that the ByteNet decoder is a state-of-the-art character-level language model based on a convolutional neural network that outperforms recurrent neural language models. We have also shown that the ByteNet generalizes the RNN Enc-Dec architecture and achieves state-of-the-art results for character-to-character machine translation and excellent results in general, while maintaining linear running time complexity. We have revealed the latent structure learnt by the ByteNet and found it to mirror the expected alignment between the tokens in the sentences.

\bibliography{main}
\bibliographystyle{icml2017}

\end{document}